\documentclass{article}

\usepackage{microtype}
\usepackage{graphicx}
\usepackage{subfigure}
\usepackage{booktabs} 

\usepackage{hyperref}

\usepackage[accepted]{mlsys2020}

\mlsystitlerunning{S3ML: A Secure Serving System for Machine Learning Inference}

\begin{document}

\twocolumn[
    \mlsystitle{S3ML: A Secure Serving System for Machine Learning Inference}



    \mlsyssetsymbol{equal}{*}

    \begin{mlsysauthorlist}
        \mlsysauthor{Junming Ma}{pku}
        \mlsysauthor{Chaofan Yu}{ant}
        \mlsysauthor{Aihui Zhou}{ant}
        \mlsysauthor{Bingzhe Wu}{pku}
        \mlsysauthor{Xibin Wu}{ant}
        \mlsysauthor{Xingyu Chen}{ant}
        \mlsysauthor{Xiangqun Chen}{pku}
        \mlsysauthor{Lei Wang}{ant}
        \mlsysauthor{Donggang Cao}{pku}
    \end{mlsysauthorlist}

    \mlsysaffiliation{pku}{Peking University, Beijing, China}
    \mlsysaffiliation{ant}{Ant Group, Hangzhou, China}

    \mlsyscorrespondingauthor{Lei Wang}{shensi.wl@antgroup.com}
    \mlsyscorrespondingauthor{Donggang Cao}{caodg@pku.edu.cn}

    \mlsyskeywords{Machine Learning, Serving System, Intel SGX}

    \vskip 0.3in

    \begin{abstract}

        We present S3ML, a secure serving system for machine learning inference in this paper.
        S3ML runs machine learning models in Intel SGX enclaves to protect users' privacy.
        S3ML designs a secure key management service to construct flexible privacy-preserving
        server clusters and proposes novel SGX-aware load balancing and scaling methods to satisfy
        users' Service-Level Objectives. We have implemented S3ML based on Kubernetes as a
        low-overhead, high-available, and scalable system. We demonstrate the system performance
        and effectiveness of S3ML through extensive experiments on a series of widely-used models.
    \end{abstract}
]



\printAffiliationsAndNotice{}  

\section{Introduction}
\label{introduction}
The breakthrough and success of machine learning (ML) in image classification, natural language processing, and many other fields have driven the recent fast growth of ML
inference services deployed in the cloud\cite{awsai,googlecloudai}, serving a large number of users from different domains. Service providers (such as mainstream cloud vendors) train ML models
offline with large datasets in advance and then deploy them online to serve users in real-time.
Normally, ML inference services are exposed to users as HTTP/RPC APIs. In such a scenario, latency is the most concerned metric for service providers,
who need to bound the service tail latency (for example, the 99 percent of request latencies are less than 500ms) to satisfy the response-time Service-Level Objectives
(SLOs)\cite{MArk}.

The satisfaction of ML inference SLOs relies on a high-available and scalable underlying serving system, which usually adopts the following distributed architecture. A
service frontend receives requests from concurrent user clients and then distributes received requests to parallel backend model server replicas. Each replica is an
instance of the same stateless server which contains the pre-trained ML model and runs in an isolated resource environment such as a Docker container. The serving system
dynamically adjusts the number of model server replicas in real-time according to users' workload to meet SLO requirements. ML serving system gains increasing concerns
from researchers in recent years\cite{Clipper,ParityModels,MArk}. These works focus on supporting heterogeneous ML frameworks, reducing money costs, mitigating service
slowdowns and failures. Unlike these existing works' attention, we present S3ML in this paper, a secure serving system that focuses on protecting users' privacy in using
ML inference services.

Security is an important topic in ML inference systems. Most ML inference systems are deployed on the top of public or private cloud infrastructures. Users need to upload
input data to the cloud datacenters to obtain inference results from service providers. However, there is no guarantee that users' data will not be stolen or abused by
service providers, infrastructure providers, or unexpected malicious attackers in this process. Thus, there is an emerging need to design an ML serving system that meets
the privacy protection requirements and some government regulations, especially for those applications where data privacy is very important (e.g., medical data analysis).
Unfortunately, security is rarely covered in the prior work on ML serving systems.

Trusted Execution Environment (TEE) is a popular solution to the trust problem between service users and service providers. Intel Software Guard Extensions (SGX)\cite{intelsgx} is a mainstream and
widely-used TEE implementation to protect the confidentiality and integrity of programs running on untrusted server platforms, first brought to the market by Intel in 2015 with the Skylake
family CPUs. Intel SGX provides a special area in memory called enclaves to host user code and data from being stolen or tampered by untrusted processes outside enclaves.
S3ML leverages Intel SGX to provide a security guarantee for ML inference service users.

Running ML inference services in Intel SGX enclaves introduces new challenges. To achieve security, S3ML encrypts data by establishing a TLS channel between the model
server and the client. The establishment of TLS relies on public key infrastructure (PKI), i.e., a certificate and a private key. The PKI is generated in the enclave,
and the private key is only visible inside the enclave during the entire process to ensure confidentiality. However, in ML serving systems, if each backend model server
replica encrypts/decrypts data with independently generated PKI, it is impossible to do load balancing and failover to ensure service high-availability and scalability.
Therefore, the first challenge is how to synchronize PKI among different backend model server enclaves and ensure this process itself is also secure and high-available.
S3ML addresses this challenge by introducing a new module called Attestation-based Enclave Configuration Service (AECS). AECS is a dedicated service that also runs inside
enclaves, responsible for generating, managing, saving, and distributing PKI. On the basis of AECS, the management of ML inference services can be significantly simplified.

The second challenge lies in that applications running in different enclaves on a host contend for a dedicated memory region called the Enclave Page Cache (EPC).
EPC has a limited size but can be over-committed with page swapping mechanism on Linux systems. Previous work reveals that EPC over-commitment would cause significant
performance degradation\cite{VAULT}. For ML inference services, this means a higher possibility of SLO violations. To overcome this challenge, S3ML uses lightweight ML
framework/models to reduce EPC consumption. Furthermore, S3ML proposes a novel strategy that considers EPC paging activities for service load balancing and scaling to
meet SLO requirements.

We have implemented S3ML based on Kubernetes\cite{k8s}, TensorFlow Lite\cite{tflite}, and Occlum\cite{Occlum}, which is a state-of-the-art LibOS for SGX. We evaluated S3ML
through extensive experiments on a series of widely-used models. Experimental results demonstrate the system performance and effectiveness of S3ML.

We make the following contributions in this paper.
\begin{itemize}
    \item To the best of our knowledge, we propose the first Intel SGX-enabled secure serving system for ML inferences.
    \item We propose a high-available architecture and mechanism for secure key management in production-grade clusters.
    \item We propose a novel SGX-aware load balancing and scaling method for SGX-enabled ML services to meet SLO requirements.
    \item We implement the system and evaluate its performance on a series of popular ML models.
\end{itemize}

\section{Background and Motivation}
\label{background}
In this section, we give a brief introduction to Intel SGX to better understand the motivation and system design of S3ML.
Date to July 2020, SGX includes two major versions, i.e., the original version, which is usually called SGX 1.0, and the latter version
with enhanced features like Enclave Dynamic Memory Management, which are called SGX 2.0. Since SGX 2.0 is supported by a limited series
of CPU families and S3ML is built on SGX 1.0, we only discuss SGX 1.0 in this paper, which already represents the core security
features of SGX.

\subsection{Intel SGX Basics}
Intel Software Guard Extensions (SGX) is an emerging technology for building Trusted Execution Environments (TEE), a popular solution to
the classical trusted computing problem of making computations securely on untrusted platforms. The SGX technique includes a set of CPU
instruction extensions and dedicated hardware architecture for providing a secure sandbox called \emph{enclave}, which protects the
confidentiality and integrity of the code and data running inside it. An enclave is encrypted automatically and decrypted only by the CPU
for the code and data running inside the enclave. SGX can protect the code and data in an enclave from being stolen or tampered by software
and hardware attacks outside the enclave. Before the initialization of an enclave, the code and data reside in the host's untrusted area.
SGX also does not have a security guarantee on the data path outside an enclave to a client. To protect secrets, a service user needs to
upload encrypted input data, which only can be decrypted inside the enclave for inferences. In S3ML, an ML inference service consists of
multiple backend model servers running in enclaves. These model servers should be equipped with the same key to implement user-transparent
load balancing and failover to achieve high-availability and scalability.

\subsection{Attestation and Sealing}
SGX provides attestation mechanisms to enable an enclave to prove that it does run the untampered code in a genuine enclave on an SGX-enabled
platform. Local Attestation proves an enclave's identity to another enclave running on the same platform, while Remote Attestation (RA) proves
an enclave's identity to a remote third-party entity.

The code and data running in enclaves are volatile. Sealing is the mechanism provided by SGX to save data to persistent untrusted storage, such
as hard drives. A sealing key is used in this process to encrypt the content in the enclave. The sealing key can only be visible inside the
enclave and is derived from a root sealing key, which is hardware-coded and unique in each Intel SGX CPU.

\subsection{Enclave Page Cache}
Enclaves run in a special memory area called the Enclave Page Cache (EPC). EPC is encrypted by dedicated hardware and is prohibited from
accessing by all non-enclave memory, including OS kernel, hypervisor, etc. EPC is divided into pages at the granularity of 4KB for allocations.
EPC has a very limited total size of 128 MB. Out of which, only about 93 MB is usable for user code and data while the rest is reserved for
enclave metadata. The EPC's size allocated to an enclave is determined at the initialization time and cannot be modified dynamically afterward.
On Linux platforms, EPC can be over-committed through the paging mechanism, which dynamically swaps pages between trusted EPC and untrusted DRAM.
EPC paging involves integrity verification that introduces an expensive overhead. Previous work shows that this overhead degrade system performance
on average by 5x \cite{VAULT}. As EPC is small and contended by all enclaves running on a machine, paging can be triggered frequently. For S3ML, it
is necessary to utilize this precious and limited EPC effectively to ensure the satisfaction of SLOs.

\subsection{Application Development}
SGX-enabled application development has two primary ways. The first is a partition-based way like using native Intel SGX SDK or frameworks like
Asylo\cite{asylo}, which divides an application into trusted and untrust partitions. Sensitive data are processed by the trusted partition running
inside enclaves. This way utilizes EPC more efficiently but requires many efforts to modify legacy non-SGX applications. Another popular way is to
use library operating systems (LibOSes)\cite{LibOS}, which run an entire application inside an enclave with system call supports. This way provides
better compatibility to legacy applications and is more friendly for developers. The existing mainstream LibOSes for SGX include Haven\cite{Haven},
SCONE\cite{SCONE}, Panoply\cite{Panoply}, Graphene-SGX\cite{Graphene-SGX}, and Occlum\cite{Occlum}. We choose Occlum to build S3ML due to its
ease-of-use, lightweight, and superior performance.

\subsection{Threat Model}
We consider the input and output data of service users for model inferences is sensitive data that needs to be protected in this paper. The model
servers and AECS run inside SGX enclaves powered by Occlum. The source code of those components running inside enclaves of S3ML is open
to service users for utilizing the SGX attestation mechanism. We assume that attackers can control both host hardware to snoop on data and host
software, including the hypervisor, OS, and other co-located applications, to perform software attacks. Denial of service attacks, side-channel
attacks\cite{Leaky,SgxPectre,Foreshadow}, covert-channel attacks on SGX are beyond the scope of this paper.

\section{System Architecture}
\label{system}
S3ML aims at meeting the urgent needs of privacy-preserving in calling ML inference services for an increasing number of users. S3ML is designed to
be a secure, high-available, and scalable system to serve ML inference workloads stably.
\begin{figure}[h]
    \centering
    \includegraphics[width=0.45\textwidth]{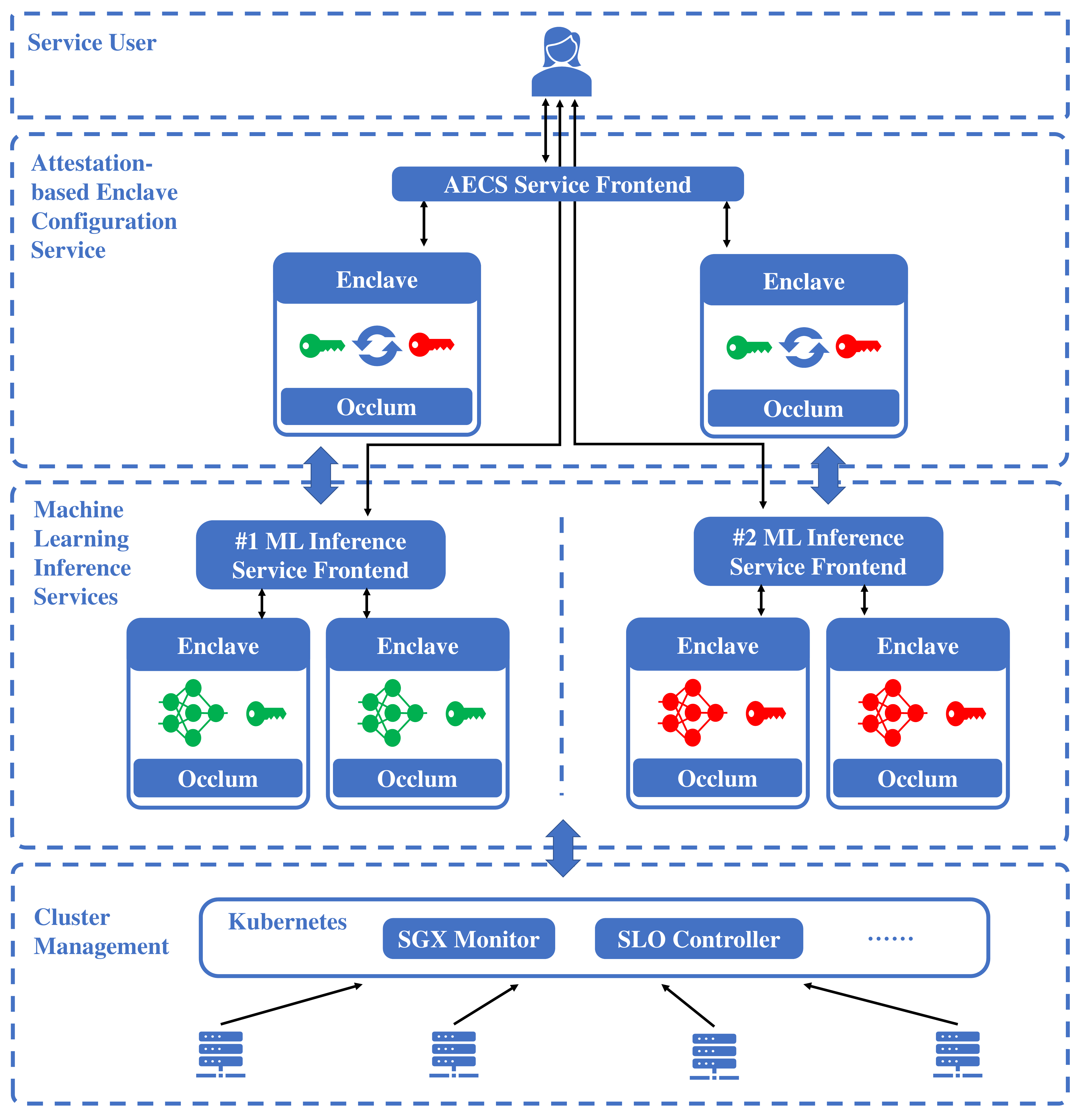}
    \caption{S3ML architecture.}
    \label{arch}
\end{figure}

Figure \ref{arch} illustrates the system architecture of S3ML. ML inference services provide users with the core functions in S3ML. Each ML inference
service in S3ML is stateless and has a cluster of backend model servers to achieve high-availability, high-throughput, and SLO requirements. Each model
server runs inside an SGX enclave. The critical point is to establish a secure communication channel between users and model servers inside the SGX
enclaves to preserve privacy when calling ML inference services. User input and output data are encrypted in the transmission path and can only be
encrypted and decrypted in the user end and inside the enclaves. For each ML service, encryption and decryption keys should be confidential and identical
for all its model servers. In S3ML, a dedicated service called Attestation-based Enclave Configuration Service (AECS) is developed for key and certificate
generation, distribution, and storage. AECS also runs inside SGX enclaves to guarantee the confidentiality of these keys. After an ML inference service
is started, each model server of the service will first contact with AECS to synchronize a certificate and a private key (see Section \ref{aecs}). The
service user also communicates with AECS to obtain the service certificate. The user and the service can then establish a secure channel to transmit
encrypted data using synchronized certificates and keys. Finally, the user sends queries to an ML inference service frontend. The frontend distributes
queries to different backend model servers according to specific load balancing algorithms.
We build S3ML based on Kubernetes, which frees us from the sophisticated cluster and service management. Furthermore, we develop add-on abilities on
Kubernetes to monitor SGX EPC activities and use it for S3ML's load balancing and scaling to satisfy SLO requirements (see Section \ref{SLO_guarantee}).



\section{ML Inference and AECS Services}
\subsection{ML inference service}
A user-facing ML inference service is exposed to users via the abstraction of Kubernetes Service at the frontend. It is mapped to a series of model servers
isolated in Kubernetes Pods at the backend. Each model server is an instance of a remote procedure call (RPC) server, running in an SGX enclave powered by
Occlum. S3ML leverages TensorFlow Lite\cite{tflite} for ML inferences to make better use of scarce SGX EPC. TensorFlow Lite is a popular ML framework
developed for edge and mobile devices with limited memory. It provides a series of size-optimized models ranging from image classification to natural language
processing. S3ML encapsulates a TensorFlow Lite interpreter into a gRPC\cite{grpc} server to implement a lightweight model server that enables users to reuse a
wide range of existing TensorFlow Lite models. Each model server is packaged into a Docker image and launched by Kubernetes to run as a Pod. All Pods of an ML
inference service belong to a Kubernetes Deployment, maintaining a stable number of running Pods by restarting the crashed ones. Thus, the high-availability of
an ML service is guaranteed.

The secure communication channel between a user and a model server is achieved via the Transport Layer Security (TLS) protocol\cite{TLS}. TLS realizes the
secure data transmission between two parties using a symmetric key exchanged through a TLS Handshake procedure. Public key infrastructure (PKI, refers explicitly
to a certificate and a private key in the rest of the paper) is needed in the TLS Handshake. S3ML does not trust digital certificates issued by third-party
certificate authorities. Instead, AECS issues self-signed certificates to services and users. The corresponding private key of each certificate is guaranteed
to be only visible inside the service enclaves. The user and the service establish an RPC over the TLS channel through the PKI generated by AECS.

\subsection{Attestation-based Enclave Configuration Service}
\label{aecs}
\begin{figure}[h]
    \centering
    \includegraphics[width=0.40\textwidth]{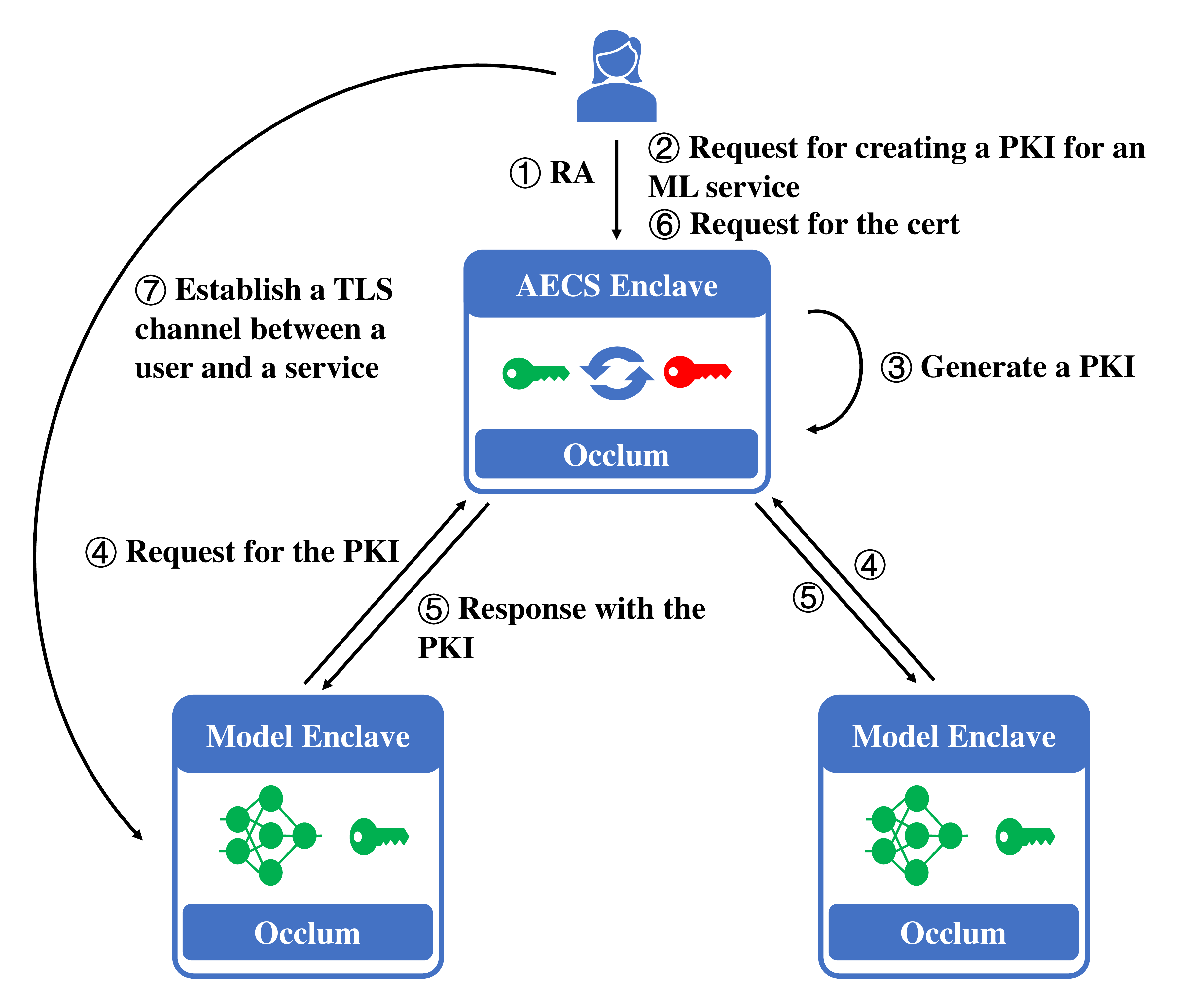}
    \caption{PKI synchronization with AECS in S3ML.}
    \label{key_sync}
\end{figure}

Like ML inference services, AECS is also exposed to users as a Kubernetes Service. AECS is the core component for key management in S3ML and consists of
multiple backend servers that are also RPC servers running inside SGX enclaves. Before calling each new ML inference service, a user needs to interact with AECS
to create a PKI for the ML inference service and obtain the certificate. Each ML model server replica also needs to first interact with AECS to obtain the PKI
to set up the TLS channel after it is started. Figure \ref{key_sync} demonstrates the process of PKI synchronization. A user first communicates with AECS and
verify AECS's identity
via RA. The user sends a request with the code hash of the model server to AECS to generate a PKI for that ML inference service. After each model server
is started, a temporary pair of public and private keys will be generated inside an SGX enclave. Each model server sends the temporary public key and its enclave
report to AECS. Next, AECS verifies each model server's identity via RA and uses the received public key to encrypt the service PKI back to each model server. Each
model server decrypts the response from AECS with the temporary private key to obtain the plain-text PKI. On the other side, the user requests AECS for the
certificate of the service. Finally, the user could establish the TLS channel with the ML inference service via the synchronized PKI. During the whole process,
the ML inference service certificate's private key is only visible inside enclaves to ensure its confidentiality.

\subsection{High-availability of AECS}
\label{aecs_ha}
Each ML inference service and user need to interact with AECS before the first call. The high-availability of other services depends on AECS, so AECS also
needs to be high-available. The high-availability of AECS is guaranteed by a Kubernetes Deployment too. However, unlike ML inference services, AECS is a
stateful service. Additional work needs to be done to achieve AECS's high-availability. AECS maintains a map that records ML inference services with their
corresponding PKIs generated inside enclaves. To synchronize this stateful map among different backend replicas, AECS encrypts this
map with a storage key and saves it to untrusted high-available persistent storage such as cloud object storage, which can be read by all backend replicas.
AECS synchronizes this storage key among all backend replicas at the time of bootstrap. Each AECS backend also saves this storage key to the local machine
storage through SGX sealing to deal with some worst scenarios such as a data center power failure that causes all AECS backends to shut down and lose the
storage key. S3ML uses the Kubernetes Selector to ensure that AECS backends are scheduled to several specific machines so that AECS can unseal the storage
key after power-on.

\begin{figure}[h]
    \centering
    \includegraphics[width=0.45\textwidth]{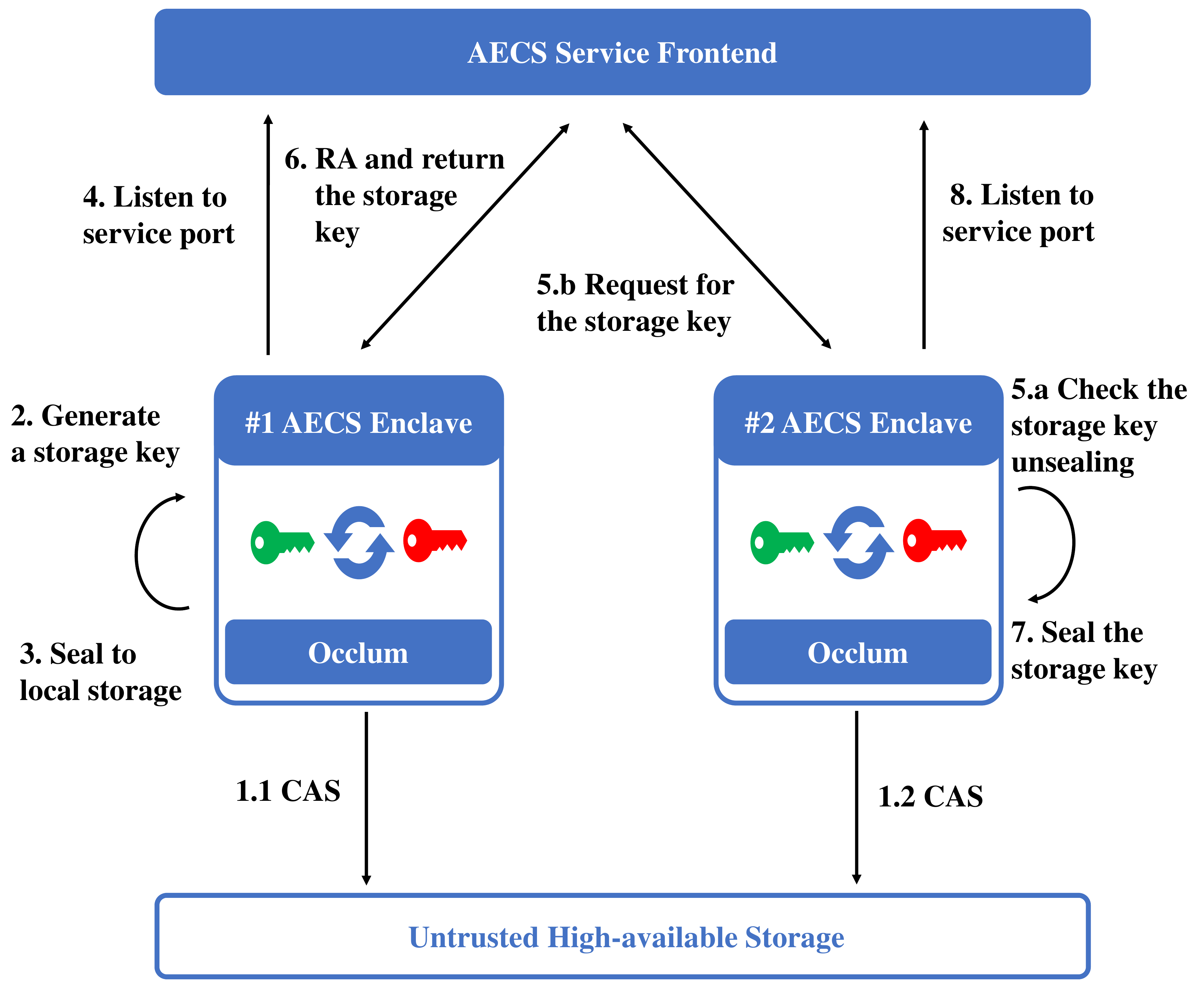}
    \caption{AECS service bootstrap.}
    \label{aecs_bootstrap}
\end{figure}

Figure \ref{aecs_bootstrap} describes the procedure of the AECS bootstrap. When an AECS backend is started, it will first determine whether this backend is the
first AECS server replica via a Compare-And-Swap (CAS)\cite{cas} operation. The first AECS server will generate a storage secret and seal it to the local storage.
The following AECS server replicas will determine whether there is sealed data on the local machine at first. If there is, the storage key can be obtained by SGX
unsealing. Otherwise, they will send their own RA report with an attached temporary public key to the AECS service.
AECS service performs RA to verify the identity of the following AECS server replicas. After the RA is finished, the storage key encrypted with the public key is
sent back to the AECS server replica. Then it obtains the plain-text storage key decrypted by the temporary private key. Finally, the AECS server replica will
start the RPC server to listen to the AECS service port. When each AECS server replica creates or deletes the PKI for an ML inference service, it will also
synchronize the encrypted map on the untrusted storage via CAS operations.

\section{SLO Guarantee in S3ML}
\label{SLO_guarantee}
In modern data centers, resource managers colocate batch jobs and latency-sensitive services to improve resource utilization\cite{Heracles}.
The existing ML inference serving systems\cite{Clipper,MArk,ParityModels} usually host backend model servers in separated Docker containers\cite{docker}.
Docker leverages \textit{cgroups}\cite{cgroup} on Linux to isolate hardware resources (i.e., CPU cores, DRAM, and I/O) for each container to
mitigate performance interference. Given a resource specification, a model server can be profiled in advance and has a predictable performance
at the runtime. Compared with traditional non-SGX applications, EPC is another special type of resource required by SGX-enabled applications.
EPC is scarce and contended by all enclaves, including system architecture enclaves on a machine. Thus, EPC paging is easily triggered to degrade
application performance, especially for latency-sensitive ML inference services. Unfortunately, there is a lack of stable and mature
\textit{cgroups}-similar mechanisms in existing operating systems to isolate EPC. In S3ML, we propose a new approach based on SGX EPC paging
activities to control network traffic and scaling to ensure the SLO of an ML inference service. We first characterize the influence of EPC
paging on service latency. Then we describe the design and implementation in S3ML to collect real-time EPC paging activities information for
load balancing and scaling.

\subsection{Service Latency and EPC Activities}
We first qualitatively observe the impact of EPC activities on service latency. In order to observe EPC paging activities, a monitor that
can provide EPC paging information is in need. However, the official SGX driver released by Intel does not have this monitor capability.
Fortunately, we can use an open-source SGX driver released by Fortanix\cite{sgxdriver}, which provides detailed EPC monitoring information,
including the number of pages resident in the EPC of each enclave, the number of pages swapped into and out from the EPC, etc.

We employ the following procedure to investigate the EPC paging impact on service latency. In the first step, we launch a single model server
on a clean SGX-enabled machine and then keep sending queries to this model server. In the second step, we run another interference program
inside an enclave on the same machine. This program requests for a continuous EPC allocation and keeps visiting consecutive pages on this
memory in an infinite loop. The purpose of this interference program is to ensure that the pages allocated to the program is always in a
\textit{fresh} state and will be resident in the EPC. As a result, the remaining EPC is left for the model server. When the remaining EPC is
insufficient for ML inferences, EPC paging will be triggered.

\begin{figure}[h]
    \centering
    \includegraphics[width=0.45\textwidth]{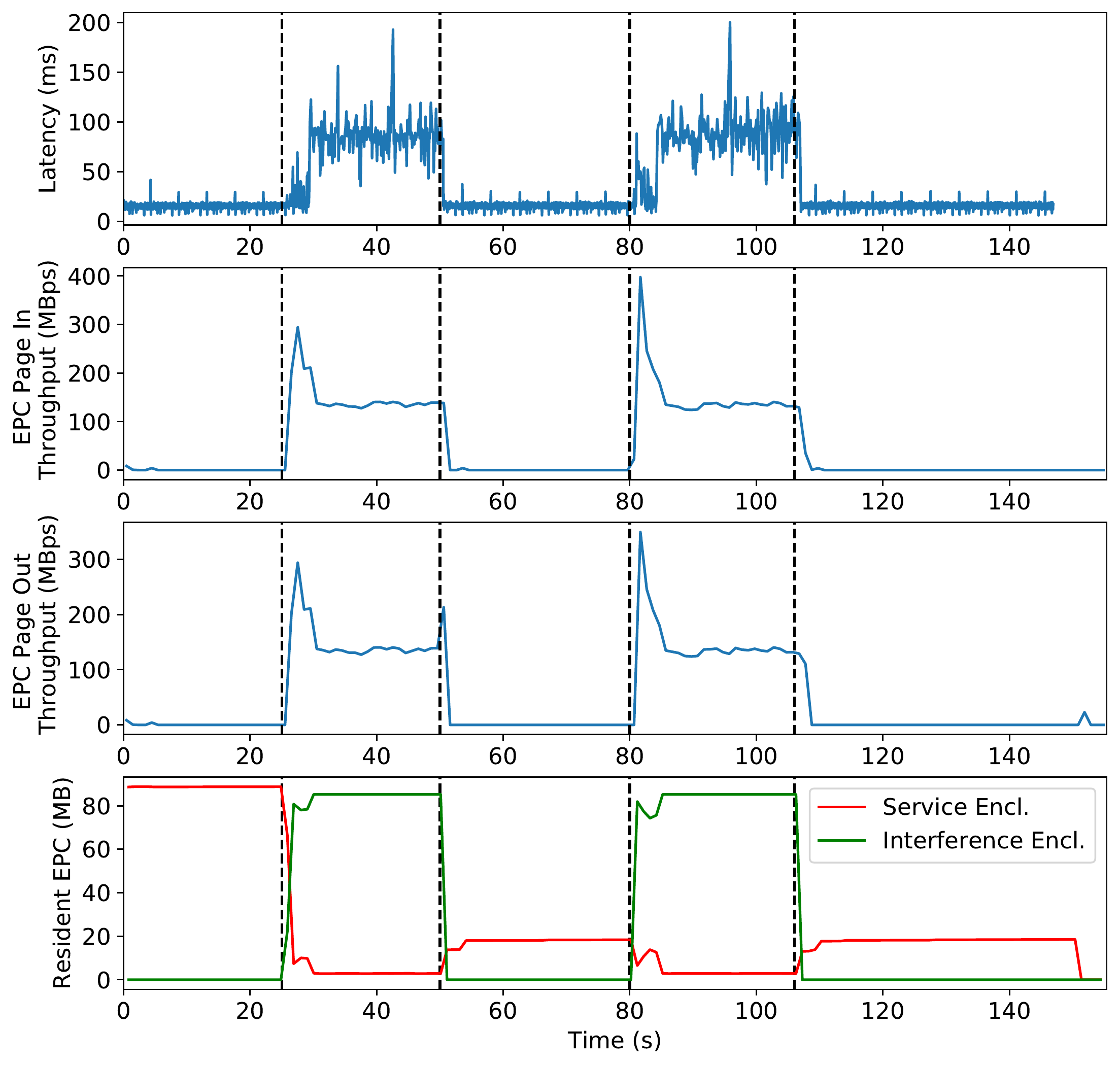}
    \caption{From top to bottom, the figures represent ML inference latency, EPC page in throughput, EPC page out throughput, and the
        resident EPC of each enclave.}
    \label{paging}
\end{figure}

We take a MobileNetV1(float)\cite{mobilenetfp} model, which does image classification tasks as an example ML inference server running
in an SGX enclave. Figure \ref{paging} shows the changes in service latency, EPC paging throughput, and resident EPC size of each enclave
during this process. When the model server started, Occlum would request more than 93 MB of EPC at the boot time, reserving to launch
other userspace processes later. Thus, the model server enclave occupies the entire EPC at the start. As the EPC allocation is determined
at the enclave initialization time, this requested EPC may not be actually used by the model server. When we launched the interference
enclave at the 25s and 80s, it preempted EPC and left the remaining EPC for the model server. It can be observed that after the
interference enclave was started, EPC pages kept being swapped in/out, which caused the increase of ML inference latency significantly.
When the interference enclave was terminated at the 50s and 106s, the EPC pages occupied by it were released, and the EPC paging
throughput dropped immediately. Then the ML inference latency becomes as low as those before the launching of the interference enclave.
In short, we can confirm that EPC paging has a significant impact on ML service latency, and we can use EPC paging activities as an
indicator of SLO violations.

\subsection{Characterizing Inference Latency under EPC Paging}
\begin{figure}[h]
    \centering
    \includegraphics[width=0.40\textwidth]{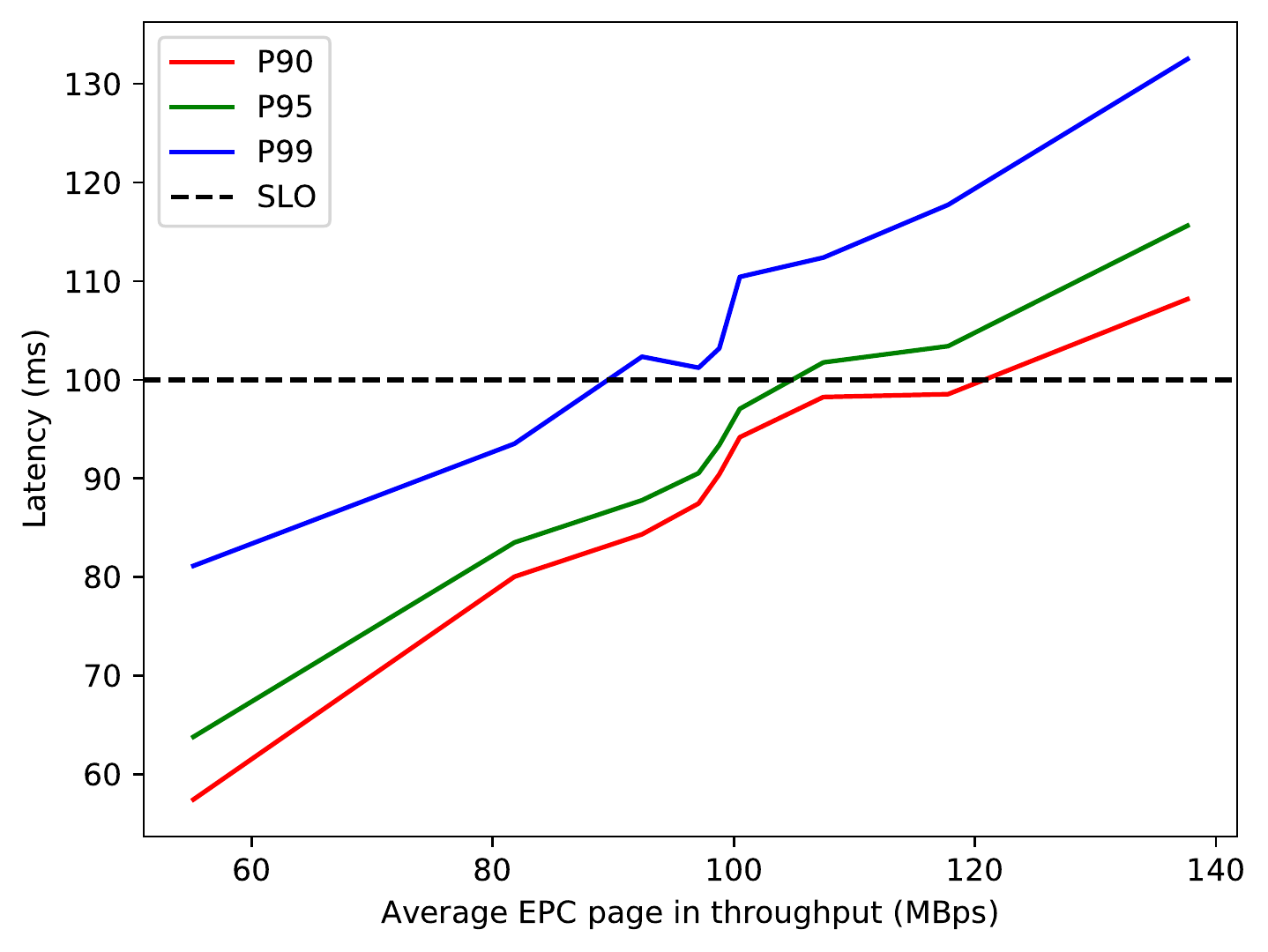}
    \caption{EPC paging influence on ML inference latency.}
    \label{paging_analysis}
\end{figure}

We quantitatively investigate the EPC paging boundary that will trigger SLO violations through off-line profiling. We firstly launch a
single model server and an interference enclave, as described in the last section. Afterward, a client keeps sending requests to the model
server. The EPC paging throughput is intentionally controlled by configuring the different EPC size requested by the interference enclave.
Figure \ref{paging_analysis} shows 90 percentile latency, 95 percentile latency, and 99 percentile latency of the MobileNetV1(float) model
serving 2k inference requests under different average EPC paging throughput. It can be seen that the EPC paging throughput and latency are
positively correlated. The higher the EPC paging throughput, the higher the latency. Given an SLO for a specific ML inference service, the
EPC paging throughput boundary that would trigger SLO violations can be determined. Note in this profiling, the pages swapped in and out of
EPC almost all belong to the model server. The reason is that we use an infinite loop in the interference enclave to refresh allocated pages
resident in the EPC. In actual scenarios with colocated batch jobs and latency-sensitive services, we can only know the total EPC paging
throughput and have no idea how many swapped pages of those belong to the model sever, which have a direct impact on the service latency.
However, using the obtained boundary as an indicator is still feasible as it is a safe value. It is based on a conservative assumption to
prevent service SLO violations because the actual latency degradation may be less than expected. Some of the swapped pages belong to other
enclave. As each ML service is different in binary size, model size, feature dimensions, computational complexity, and SLO requirements,
individual off-line profiling is needed to determine each ML service's SLO violation boundary.

\subsection{Load Balancing}
\subsubsection{Control Logic}
\label{lb-logic}
To mitigate the impact of EPC paging on ML inference latency, we propose an SGX-aware load balancing in S3ML. Our SGX-aware load balancing
extends from the existing Kubernetes load balancing algorithms. We do not directly decide how to distribute requests from the frontend to the
backend. Instead, we control whether a backend model server replica is eligible to serve requests according to real-time EPC activities. Our
load balancing control logic follows a threshold-based reactive procedure. S3ML periodically monitors the EPC activities on the corresponding
cluster node of each enclave. Suppose the EPC paging throughput (both page in and out) of one node is higher than a threshold, i.e., a specific
percentage of the SLO violation boundary(such as 70\%) for several consecutive monitor cycles. In that case, S3ML stops the network traffic
from the service frontend to the backend model server replica. When S3ML detects no more interference enclaves (except system architecture
enclaves) running on the node, S3ML resumes the network traffic to the model server replica to serve users again.

\subsubsection{Implementation}
We implemented SGX-aware load balancing based on Kubernetes Service.

\textbf{EPC information monitoring.} We use a popular open-source monitoring framework on Kubernetes, i.e., Prometheus\cite{prometheus}, to
collect the EPC activity information on each node. Prometheus periodically pulls information such as hardware resource usage on each node
through a daemon called \textit{node exporter} and gathers them together. However, the current node exporter does not support to collect the
information on SGX EPC activities. Thus, we have implemented a new collector for SGX EPC resources in the Prometheus node exporter. This
collector collects SGX EPC usage information including the total number of swapped pages and running enclaves' resident EPC by reading the
\textit{/proc/sgx\_enclaves} and \textit{/proc/sgx\_stats} files exposed by the Fortanix SGX driver\cite{sgxdriver}.

\textbf{Load balance control.} Our SGX-aware load balancing is extended from the native Kubernetes IPVS-based service load balancing\cite{k8ssvc},
which is the most advanced Layer 4 load balancing functionality in the native Kubernetes. This functionality is achieved by a component called
\textit{kube-proxy} on each node, which sets a series of IPVS\cite{lvs} forwarding rules to distribute network traffic from a service frontend
to different backend model server Pods. The IPVS-moded kube-proxy currently supports six IPVS scheduling algorithms, including Round-Robin(RR),
Least Connection(LC), Shortest Expected Delay(SED), and etc\cite{lvsalgs}. No matter which scheduling algorithm is chosen, kube-proxy sets all
backend model servers with the same IPVS weight (hard-coded as 1 in source code), which means all model servers are considered to have fixed
processing capability.

In S3ML, we achieve SGX-aware load balancing via dynamically changing the IPVS weights according to real-time EPC activities based on the native
SED algorithm. We implemented a custom Kubernetes Controller called SLO Controller, which periodically obtains each node's EPC activity
information from Prometheus and the backend endpoints of each service from the Kubernetes API Server. The controller follows the procedure described
in section \ref{lb-logic}, setting the weight of a backend to 0 (i.e., no traffic is allowed) when consecutive intense EPC paging activities are
detected and setting its weight back to 1 when no other interference enclaves are detected. Finally, we modify the kube-proxy to enable it to update
the underlying IPVS weights on each node.

\subsection{Scaling}
Scaling is an essential feature for a serving system to achieve high-throughput and meet SLO requirements under dynamic workload arrivals and
unexpected EPC interferences. Many works\cite{scaling1,scaling2,scaling3} have been done on the improvement of autoscaling algorithms. These
techniques achieve effective scaling through reactive rule-based methods or proactive methods based on time-series predictions.

In this paper, we do not innovate on the front of scaling algorithms. The scaling of S3ML is based on the Kubernetes Horizontal Pod Autoscaling\cite{k8shpa},
which supports to create and delete Pods dynamically to maintain the stability of custom metrics. S3ML takes the aggregate CPU utilization of the
model server replicas, which are in actual service (i.e., load balancing weight is not 0) as the metric. In this way, scalability can be achieved
based on the actual serving capacity of S3ML. We also set this metric to reserve some resources for burst requests and sudden enclave interferences.

\begin{figure*}[h]
    \centering
    \subfigure[MobileNetV1(float)]{
        \label{fig:a}
        \includegraphics[width=0.22\textwidth]{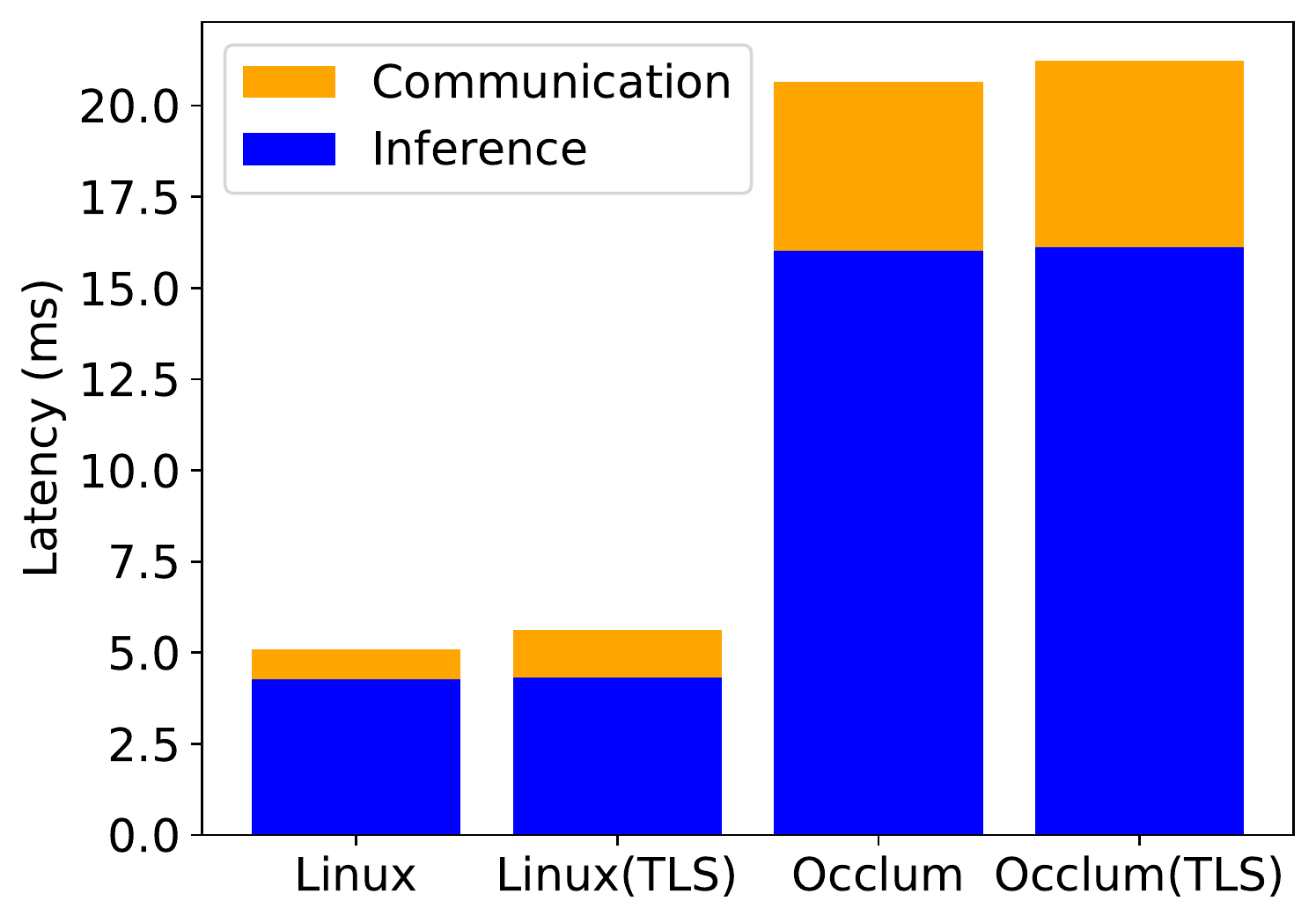}
    }
    \subfigure[MobileNetV1(quantized)]{
        \label{fig:b}
        \includegraphics[width=0.22\textwidth]{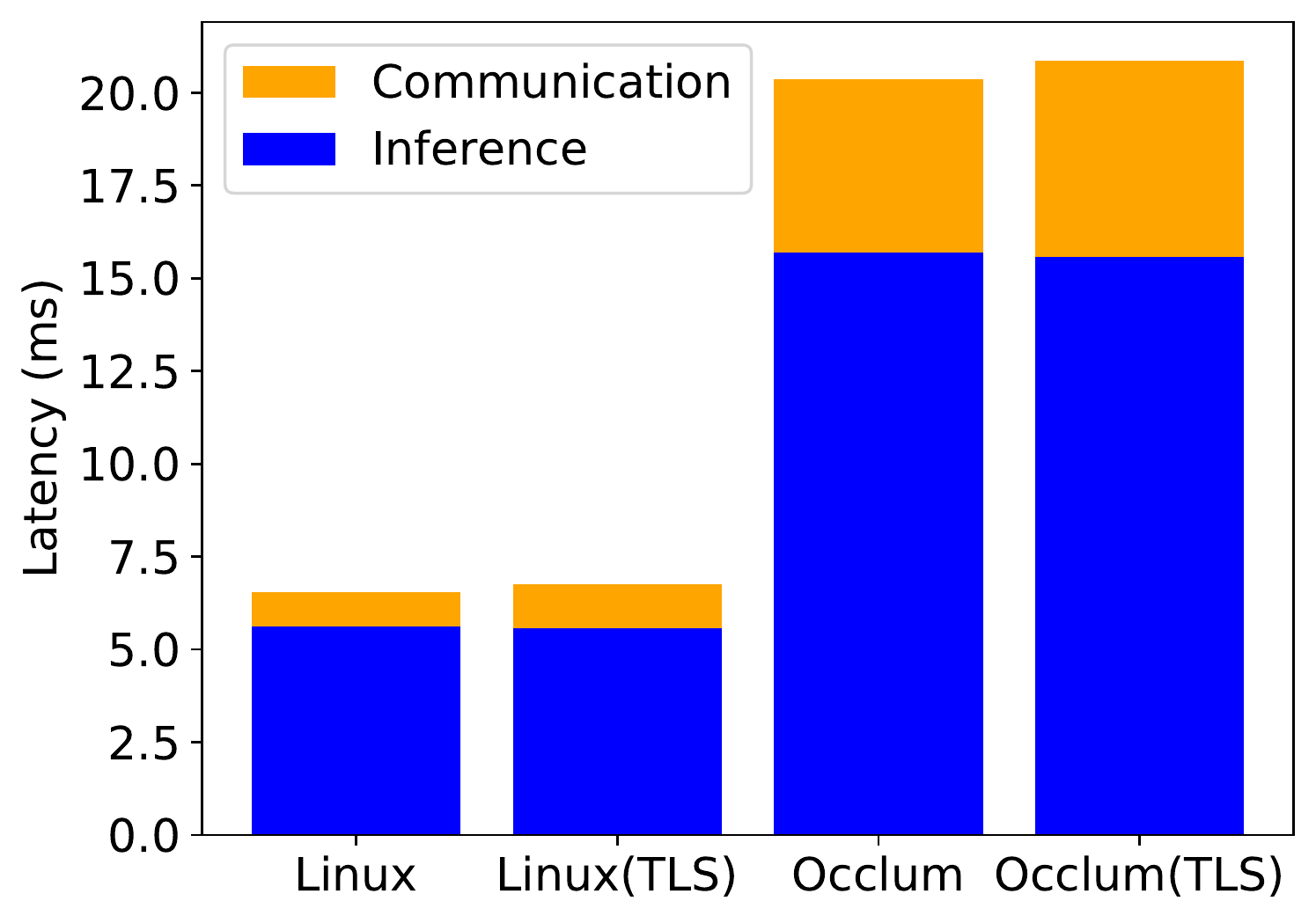}
    }
    \subfigure[EfficientNetLite(float)]{
        \label{fig:c}
        \includegraphics[width=0.22\textwidth]{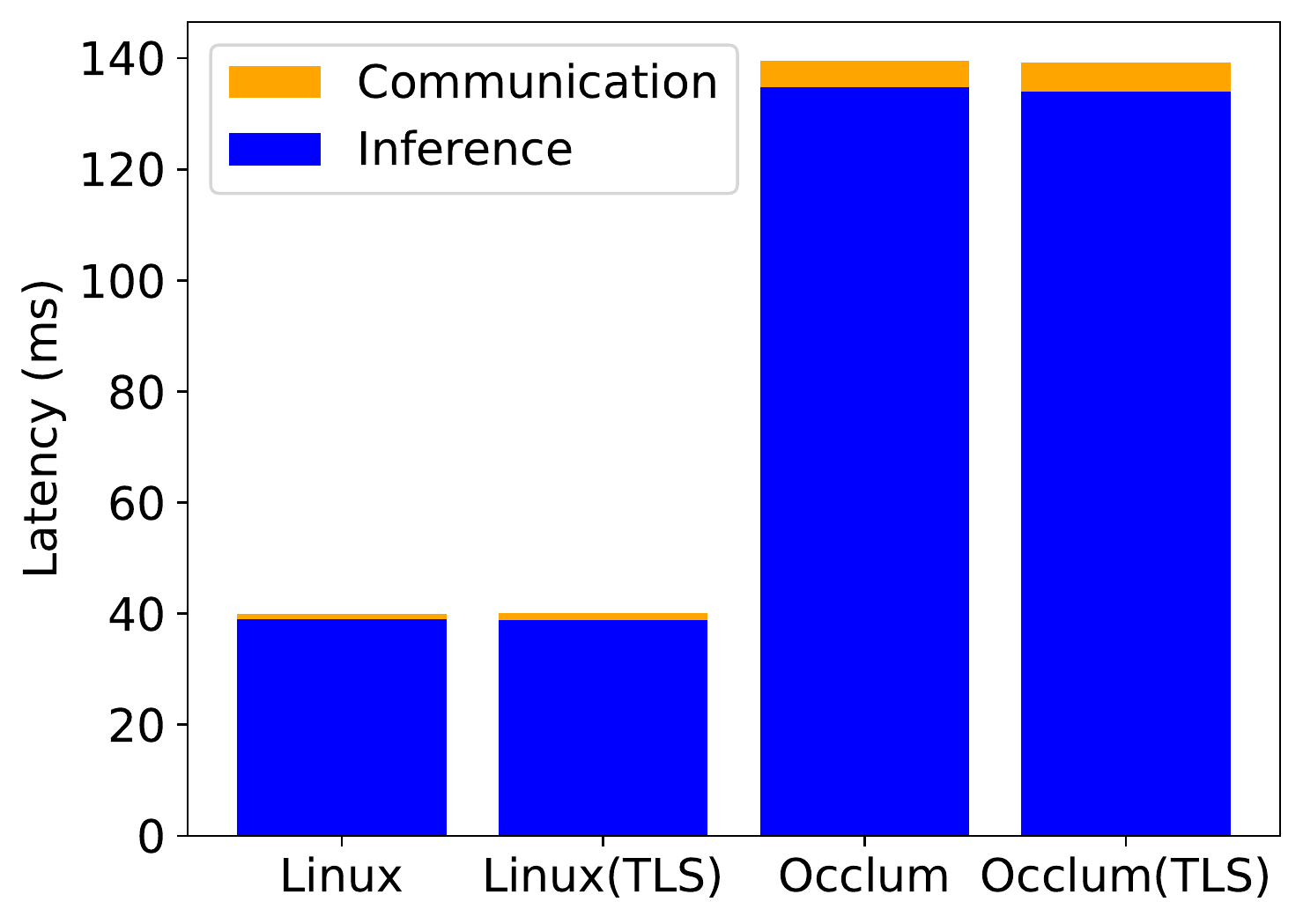}
    }
    \subfigure[EfficientNetLite(quantized)]{
        \label{fig:d}
        \includegraphics[width=0.22\textwidth]{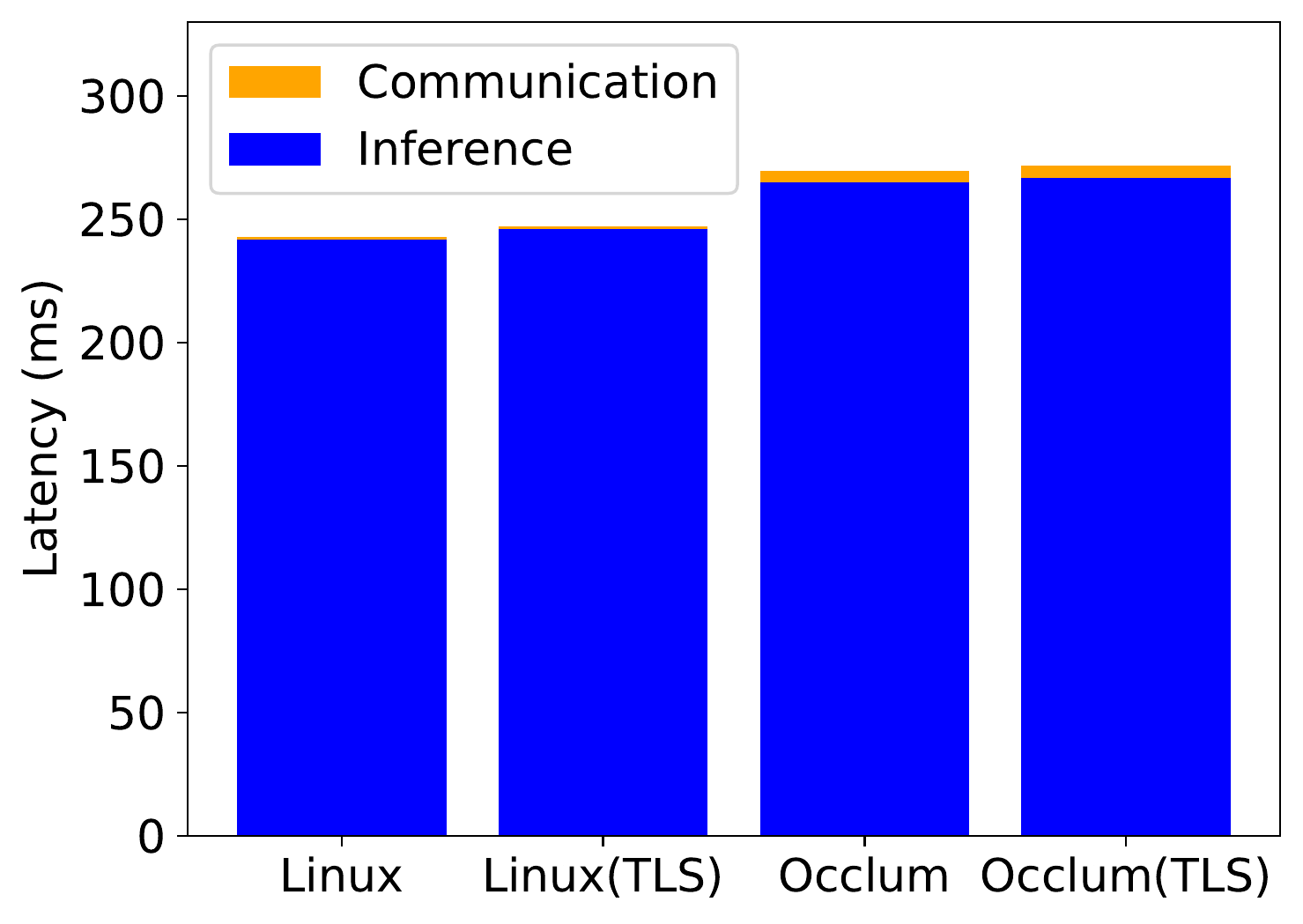}
    }
    \caption{Inference and communication latency of four popular lightweight ML models. From left to right in each figure, the bars stand for
        running model servers in Linux without TLS, Linux with TLS, Occlum without TLS, and Occlum with TLS, respectively. The blue bars represent
        the time used for inferences, and the orange bars represent the time for communication between the client and the server.}
    \label{performance}
\end{figure*}

\section{Evaluation}
\label{evaluation}
In this section, we present the evaluation of S3ML on system performance, load balancing effectiveness, and scalability using a series of representative ML
models. All the experiments are done on the Alibaba Cloud\cite{aliyun}. Firstly, we compare the system performance of S3ML to serving systems in non-SGX
environments with/without TLS communications. After that, we compare S3ML's SGX-aware load balancing to the native Kubernetes load balancing algorithms. Finally,
we evaluate the scalability of S3ML.

\subsection{Experimental Setup}
\label{expsetup}
\textbf{Testbed.}
We build a three-node Kubernetes cluster with three Alibaba Cloud \textit{ecs.ebmhfg5.2xlarge} bare-metal instances. The Kubernetes master node is configured
to enable Pods to be scheduled on it. Each instance has 8 CPUs, 32 GB of RAM, and an SGX-enabled Intel Xeon E3-1240v6 (Skylake) CPU. All instances are connected
via a high-bandwidth internal network. S3ML runs on CentOS 7.2, Linux Kernel 3.10.0-514.6.2.el7.x86\_64, Docker 19.03.12, and Kubernetes 1.18.5. S3ML services
are developed based on gRPC 1.26.0, TensorFlow Lite 1.15.0, and Occlum 0.15.0.

\textbf{Inference task and ML models.}
We use the image classification task to evaluate S3ML. Image classification is one of the most representative machine learning tasks. It has a wide range of
application scenarios and has made many breakthroughs in recent years. In the image classification task, the input image and the output result are regarded as
sensitive data that we need to protect. We use four models, i.e. MobileNetV1(float)\cite{mobilenetfp}, MobileNetV1(quantized)\cite{mobilenetquant},
EfficientNetLite(float), and EfficientNetLite(quantized)\cite{efficientnet} covering different model sizes and inference latencies
\footnote{Particularlly, the models used for evaluation are labeled as "mobilenet\_v1\_0.25\_128", "mobilenet\_v1\_0.25\_128\_quantized", "efficientnet/lite0/fp32", and "efficientnet/lite0/int8" respectively on TensorFlow Hub.}
in evaluation. Both MobileNetV1 and EfficientNetLite are representative and advanced image classification models, optimized for small memory devices, and trained
on the ImageNet(ILSVRC-2012-CLS)\cite{ILSVRC15} dataset. Float means that it is a floating-point model, while quantized means that it is a quantized model. The
queries are also drawn from the ImageNet dataset for evaluation.

\subsection{System Performance}
\textbf{Setting.}
In this experiment, we have four environment settings, i.e., Linux without TLS, Linux with TLS, Occlum without TLS, and Occlum with TLS, for ML inferences. The
Linux environments mean not using SGX, while the Occlum environments mean running model servers in SGX enclaves. The Occlum with TLS is what we actually used in
S3ML. The other three settings are evaluated for comparisons. For each setting, we launch a single model server. TensorFlow Lite interpreter is tuned to use four
threads.

\textbf{Result.}
Figure \ref{performance} reports the average inference latency and the average communication latency of the four models serving 10k inference requests under the
four settings. It can be found that running model servers in SGX enclaves powered by Occlum introduces additional inference and communication overhead compared
with running in Linux. For inference latency, it causes 1.09$\sim$3.77x slowdown, which depends on what model is used. The communication overhead is model-independent
and is up to the input and output data. The slowdown is 4.64$\sim$5.78x in the TLS-disabled setting and 3.90$\sim$4.48x in the TLS-enabled setting.
Furthermore, using TLS is more time-consuming than plain-text communication due to the additional data encryption and decryption in transmission. However, compared
to ML inferences' total response time, this overhead is negligible, especially for slow models.

S3ML establishes TLS secure channels between a model server running in the SGX enclave and clients to run ML inferences. The response time lies in the order of tens
of milliseconds to hundreds of milliseconds. For user-facing ML inference services, this performance overhead is acceptable.

\begin{table*}[t]
    \caption{The 99 percentile latency of four load balancing algorithms of four models under high-degree and low-degree interference settings.}
    \label{lb-table}
    \vskip 0.15in
    \begin{center}
        \begin{small}
            \begin{sc}
                \begin{tabular}{l|c|c|c|c|c|c|c|r}
                    \toprule
                    interference degree & \multicolumn{4}{c|}{High}                       & \multicolumn{4}{c}{Low}                                                             \\
                    \midrule
                    Model Name          & \multicolumn{8}{c}{MobileNetV1(float)}                                                                                                \\
                    \midrule
                    Algorithm           & RR                                              & LC                      & SED    & SGX-aware & RR     & LC     & SED    & SGX-aware \\
                    P99 latency (ms)    & 815.36                                          & 683.63                  & 101.11 & 35.92     & 25.20  & 32.50  & 30.46  & 30.55     \\
                    \midrule
                    Model Name          & \multicolumn{8}{c}{MobileNetV1(quantized)}                                                                                            \\
                    \midrule
                    Algorithm           & RR                                              & LC                      & SED    & SGX-aware & RR     & LC     & SED    & SGX-aware \\
                    P99 latency (ms)    & 733.13                                          & 568.85                  & 74.61  & 35.58     & 25.53  & 31.44  & 29.08  & 29.74     \\
                    \midrule
                    Model Name          & \multicolumn{8}{c}{EfficientNetLite(float)}                                                                                           \\
                    \midrule
                    Algorithm           & RR                                              & LC                      & SED    & SGX-aware & RR     & LC     & SED    & SGX-aware \\
                    P99 latency (ms)    & 3059.76                                         & 527.17                  & 379.09 & 217.50    & 188.69 & 163.17 & 157.05 & 158.89    \\
                    \midrule
                    Model Name          & \multicolumn{8}{c}{EfficientNetLite(quantized)}                                                                                       \\
                    \midrule
                    Algorithm           & RR                                              & LC                      & SED    & SGX-aware & RR     & LC     & SED    & SGX-aware \\
                    P99 latency (ms)    & 3863.17                                         & 552.40                  & 540.10 & 429.48    & 394.88 & 393.61 & 391.14 & 395.13    \\

                    \bottomrule
                \end{tabular}
            \end{sc}
        \end{small}
    \end{center}
    \vskip -0.1in
\end{table*}

\subsection{Load Balance}
\textbf{Setting.}
In this experiment, we evaluate S3ML's SGX-aware load balancing ability to mitigate interference from other co-located enclaves. We create an ML inference
service on the Kubernetes cluster with three backend model server replicas. Each node hosts one of the replicas. A client sends inference queries, whose
arrival time follows the Poisson distribution. For different models, we control the query rate by configuring different mean values of the Poisson
distribution. In each run, the client keeps sending queries for 10 minutes. During two time periods of 2-4 minutes and 6-8 minutes, we intentionally launch
a batch task in an SGX enclave as an interference task. As there are no publicly available SGX application datasets for evaluation currently, we run the
Stress-SGX\cite{stresssgx} program, an open-source SGX testing program, as the interference batch task. To demonstrate that SGX-aware load balancing can
handle EPC interference activities with different degrees properly, we conduct two tests for each model and each algorithm. We allocate different EPC sizes
for the Stress-SGX program to simulate high-degree EPC interference and low-degree EPC interference in each test. High-degree interference is defined as the
program would keep causing paging throughput exceeding the control threshold. In contrast, low-degree interference will not cause continuous intense EPC
paging activities. As the interference SGX enclave may lead to the degradation of system serving
capacity, the query rate is set to be fulfillable even with the remaining replicas.

For S3ML SGX-aware load balancing, the EPC monitoring information is obtained once per second. The consecutive monitoring cycles and the percentage to trigger
load balancing weight updates are tuned as five and 70\%.

\begin{figure*}[h]
    \centering
    \subfigure[MobileNetV1(float)]{
        \label{fig1:a}
        \includegraphics[width=0.22\textwidth]{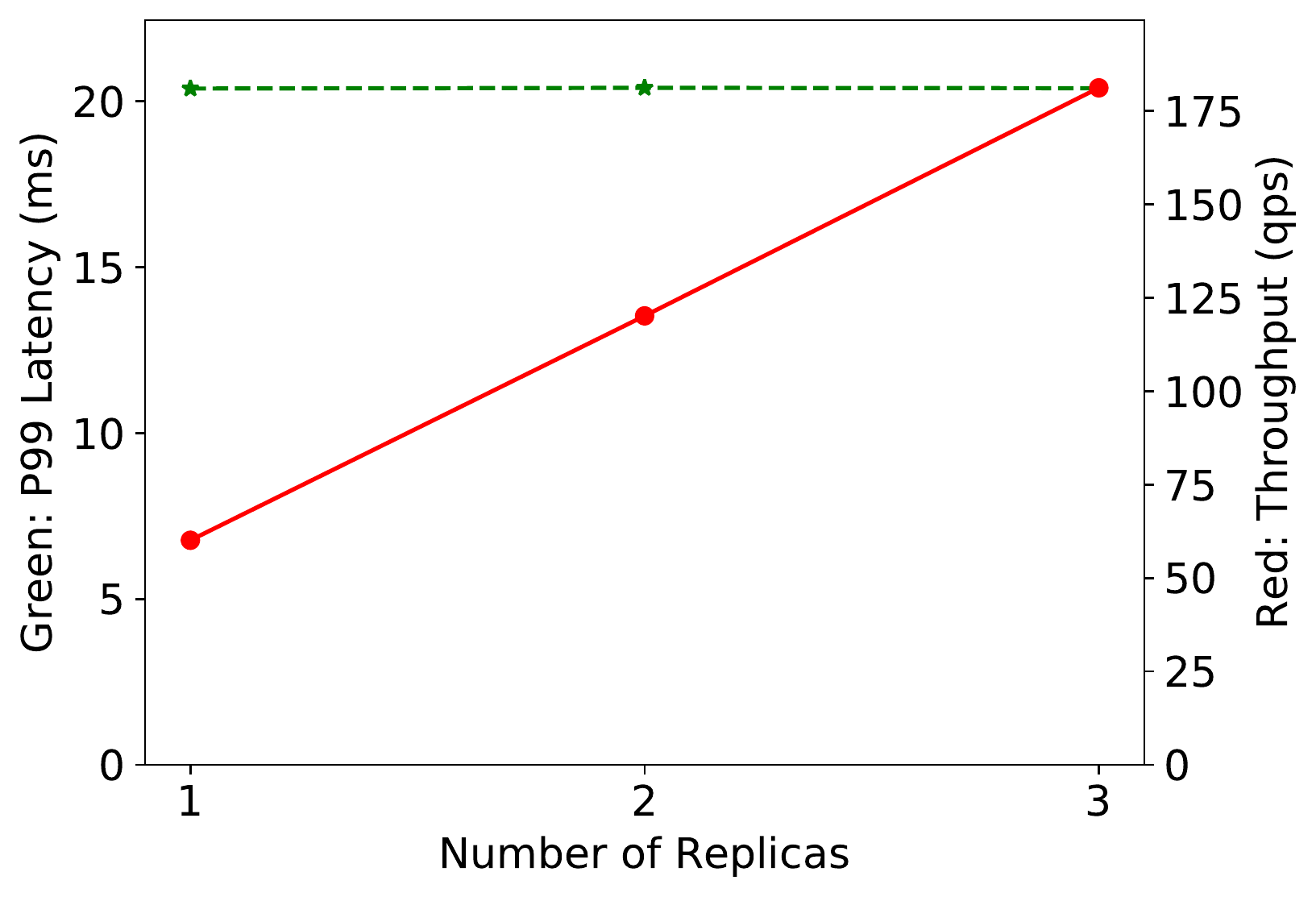}
    }
    \subfigure[MobileNetV1(quantized)]{
        \label{fig1:b}
        \includegraphics[width=0.22\textwidth]{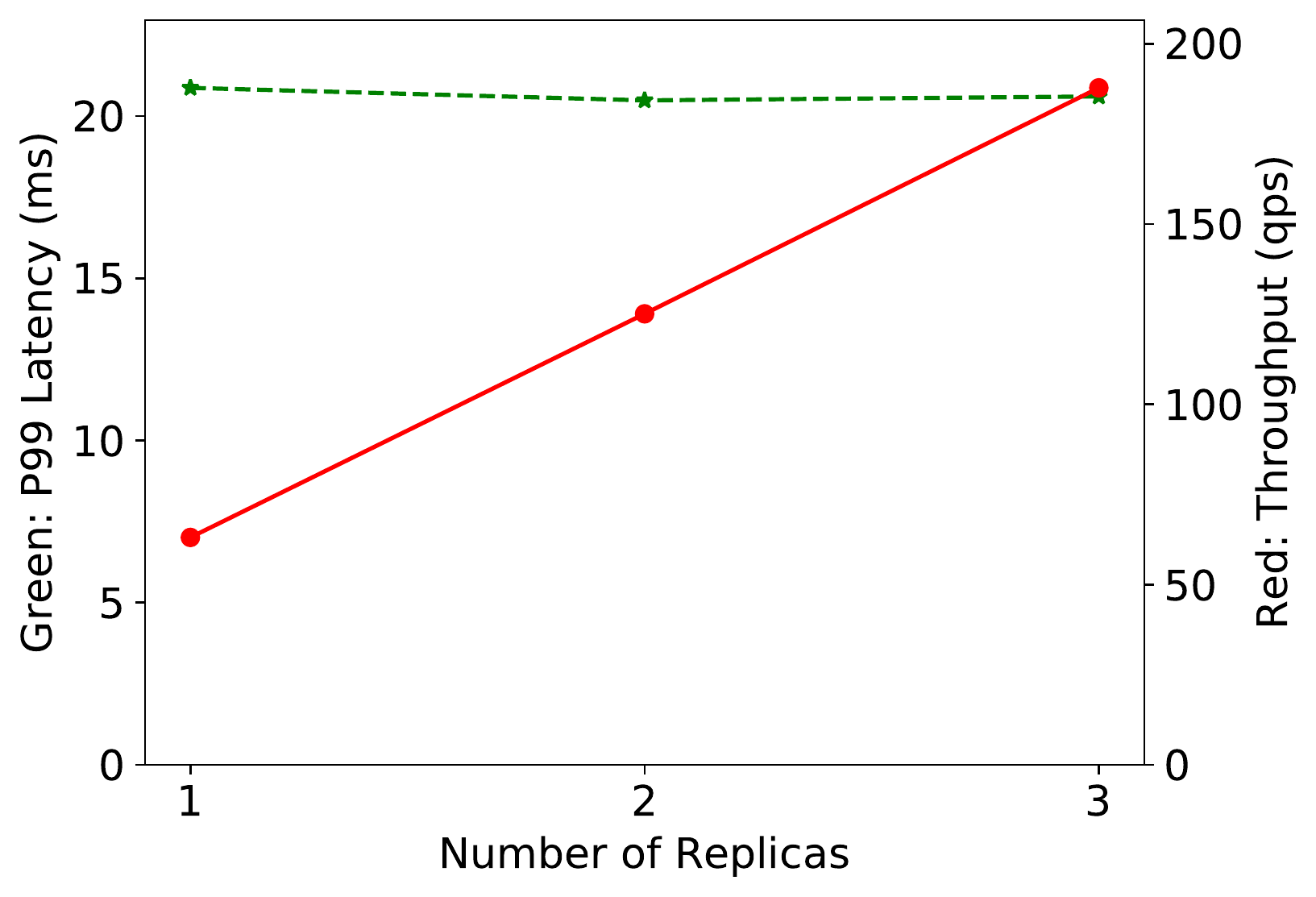}
    }
    \subfigure[EfficientNetLite(float)]{
        \label{fig1:c}
        \includegraphics[width=0.22\textwidth]{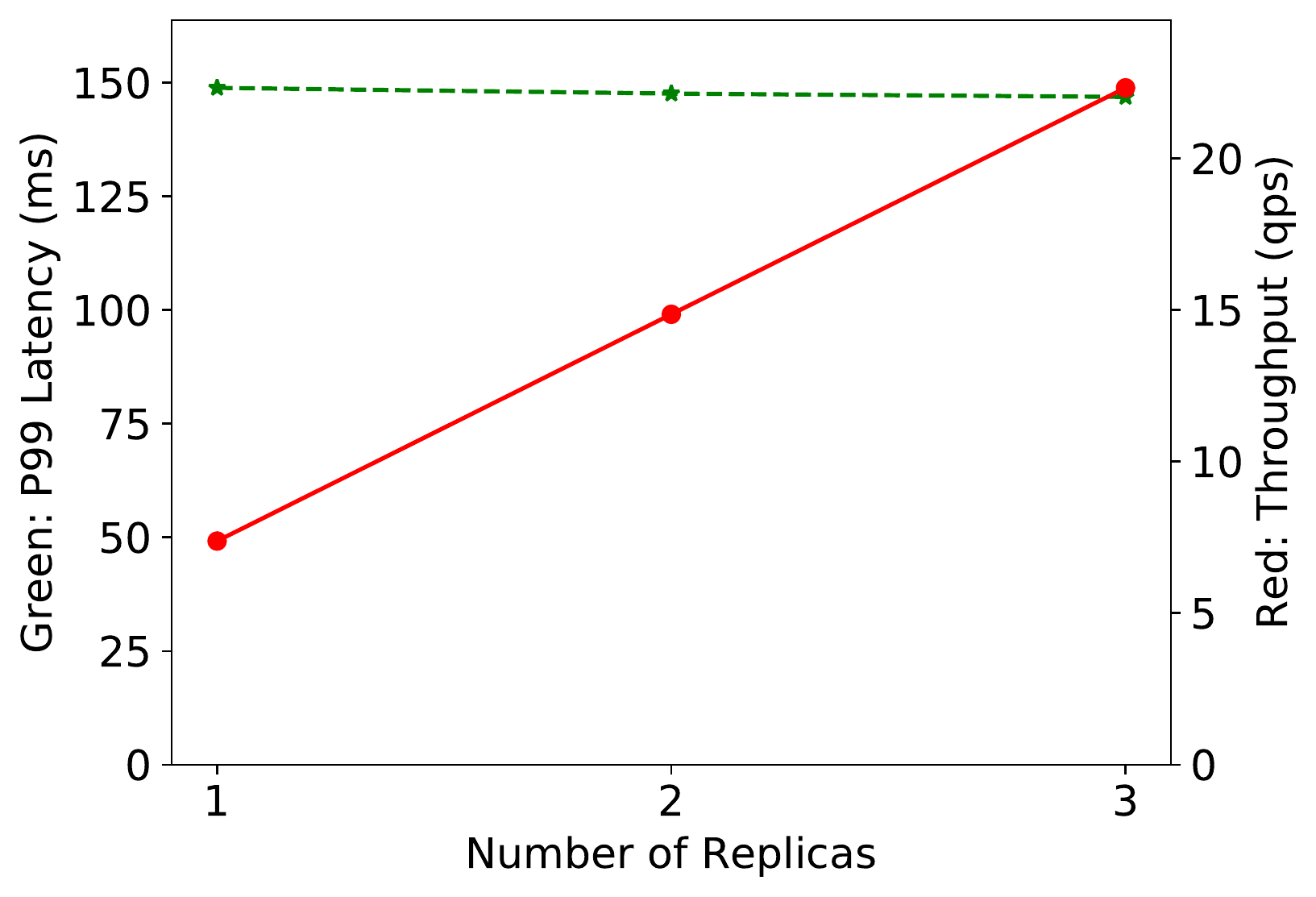}
    }
    \subfigure[EfficientNetLite(quantized)]{
        \label{fig1:d}
        \includegraphics[width=0.22\textwidth]{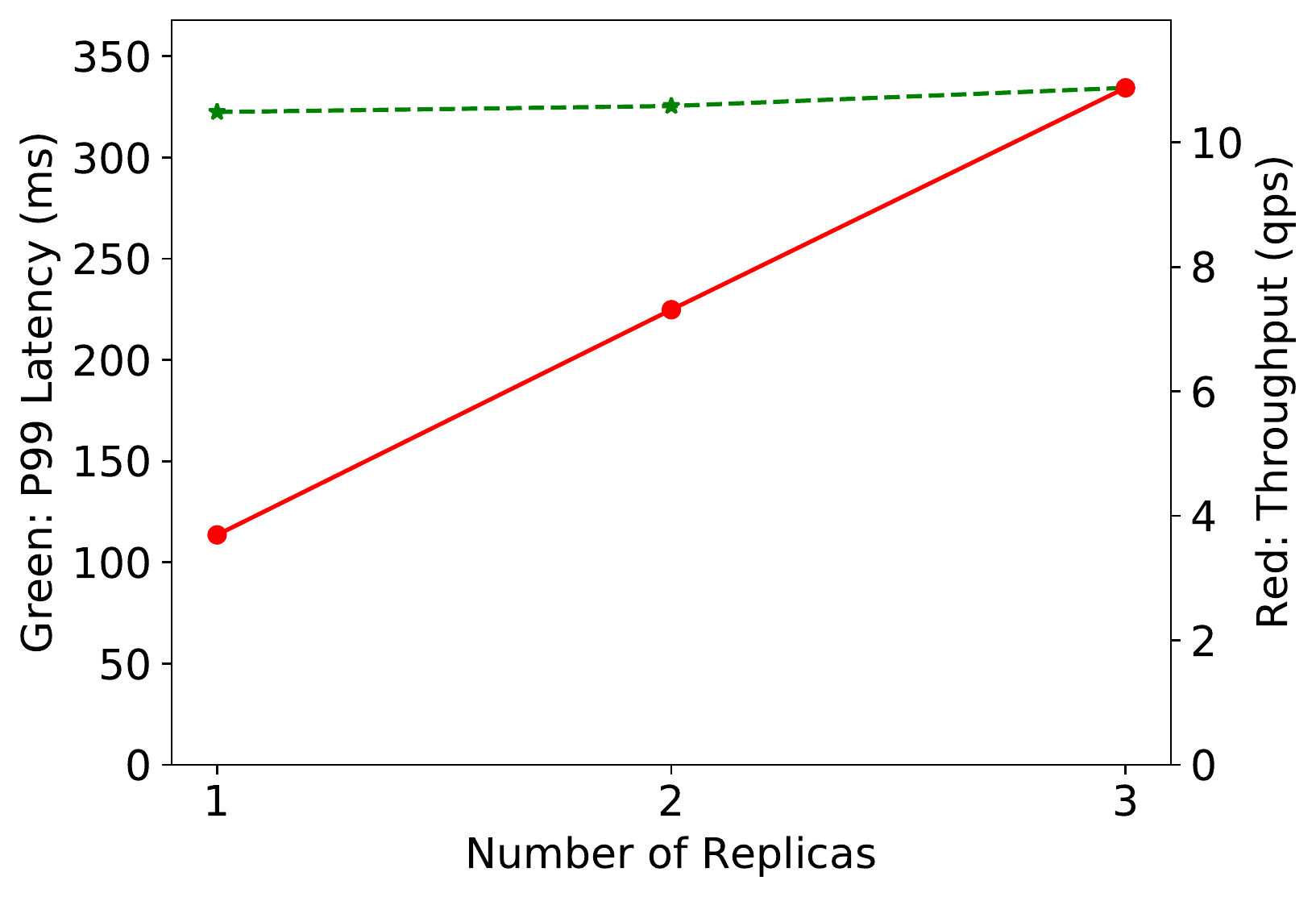}
    }
    \caption{The 99 percentile latency and throughput of four models with different numbers of replicas in S3ML.}
    \label{scaling}
\end{figure*}

\textbf{SLO.}
Latency measures the time between when a model server receives a request and when the model server returns the response. We use the 99 percentile tail latency
as the SLO metric. For MobileNetV1(float), MobileNetV1(quantized), EfficientNetLite(float), and EfficientNetLite(quantized), the SLOs are set as 100ms, 100ms,
500ms, and 600ms.

\textbf{Baseline.}
We compare SGX-aware load balancing with three native Kubernetes load balancing algorithms, i.e., Round-Robin (RR), Least-Connection (LC), and Shortest Expected Delay (SED).
\begin{itemize}
    \item RR: the arriving request is distributed to the next server.
    \item LC: the arriving request is distributed to the server with the least active connections.
    \item SED: the arriving request is distributed to the server with the shortest expected delay.
\end{itemize}

\textbf{Result.}
Table \ref{lb-table} reports the 99 percentile latency comparison of four models using four different load balancing algorithms under high-degree and low-degree
interference conditions. In the case of high-degree interference, among the three native Kubernetes load balancing algorithms, RR has the worst performance. The
reason is that when a high-degree EPC interference occurs on a node, the inference latency on the co-located model server will increase. However, the RR algorithm
still keeps distributing arriving requests to that model server. The model server will then be overloaded, which makes more requests have long latencies, eventually
leading to SLO violations. In contrast, SED has the best performance out of the native algorithms as it will distribute requests to servers with low expected latency
as much as possible. Our SGX-aware algorithm obtains the best performance among all the four load balancing algorithms. It is the only one that satisfies all the
SLOs of the four models. Specifically, for each model, compared to the other three algorithms, the SGX-aware algorithm reduces the 99 percentile latency by
64.47\%$\sim$95.59\%, 52.31\%$\sim$95.15\%, 42.63\%$\sim$92.89\%, and 20.48\%$\sim$88.88\% respectively.

In the low-degree interference setting, none of the four load balancing algorithms cause SLO violations, and there is no significant difference in performance. The
results demonstrate that our SGX-aware algorithm can effectively deal with different degrees of EPC interference.

\subsection{Scalability}
\label{scalability}

\textbf{Setting.}
We evaluate the scalability of S3ML in this experiment. We use a different number of model server replicas in each round to serve inference requests for each model.
Each replica is scheduled on an individual node.

\textbf{Result.}
Figure \ref{scaling} gives the 99 percentile latency and throughput of four models with different numbers of replicas. The tail latency shows no significant
differences between single and multiple replicas. The throughput shows a linear relationship with the number of replicas. Compared to a single replica,
S3ML increases the throughput with three replicas by 3.01x, 2.98x, 3.03x, and 2.94x, respectively, for each model. The results demonstrate S3ML's linear scalability.

\section{Related Work}
\label{related_work}

We survey the related work on ML serving systems and ML frameworks in TEE.

\subsection{Serving System}
Clipper\cite{Clipper} is a low-latency layered serving system. By dividing a serving system into a model selection layer and a model abstraction layer, Clipper
achieves the integration of different ML frameworks (such as Spark\cite{spark}, TensorFlow\cite{tensorflow}) in one system. Furthermore, caching, batching, and
other techniques are used to reduce inference latency and improve inference throughput and accuracy. MArk\cite{MArk} and Swayam\cite{Swayam} are cloud-based
serving systems designed to achieve the dual goal of reducing the money costs while satisfying SLOs. ParM\cite{ParityModels} proposes a novel learning-based
approach for enabling erasure-coded resilience in serving systems. TensorFlow-Serving\cite{tfserving} is a production-grade serving system for ML with
high-performance and flexible model version management. Rafiki\cite{rafiki} provides both ML training and inference services. It uses online ensemble modeling
to make a tradeoff between inference latency and accuracy. S3ML is orthogonal to the existing serving systems. The focus of S3ML is to provide privacy-preserving
ML inference services to users with Intel SGX. Furthermore, S3ML proposes SGX-aware SLO guarantee approaches to serve users with low latencies.

\subsection{Machine learning frameworks in TEE}
Recently, there have been a series of works using TEE to preserve privacy in data processing systems\cite{VC3,Opaque} and storage systems\cite{EnclaveDB,Pesos}.
In ML, TensorSCONE\cite{TensorSCONE} implements a secure ML system by integrating TensorFlow with SCONE\cite{SCONE} to running inside SGX enclaves. SLALOM\cite{Slalom} divides
the computations of deep neural networks (DNN) into trusted and untrusted partitions to achieve a secure and high-performance DNN computations based on SGX.
Comparing these works to S3ML, they mainly focus on using SGX for secure ML training while S3ML focuses on secure ML serving. The most relevant work with S3ML is
TF Trusted\cite{tftrusted}, which achieves confidential ML inferences via running ML models inside SGX enclaves based on Asylo\cite{asylo} and TensorFlow Lite\cite{tflite}.
Compared with TF Trusted, S3ML builds a more complicated distributed ML serving system to provide secure, high-available, scalable, and low-latency inference services
for users.

\section{Conclusion}
\label{conclusion}
In this paper, we propose S3ML, which is a secure serving system for ML inference. S3ML targets at solving the privacy-preserving problem in
using ML inference services. S3ML leverages Intel SGX to protect the confidentiality and integrity of user data. By designing a secure key
management service (AECS), secure model server clusters are easy to be developed to build high-available and scalable ML inference services. Furthermore,
we propose to use real-time SGX EPC activities for S3ML service load balancing and scaling to meet users' SLO under the circumstance of contended EPC using.
Through extensive experiments with a series of popular models, we demonstrate the system performance and effectiveness of S3ML.



\clearpage
\nocite{langley00}

\bibliography{ref}
\bibliographystyle{mlsys2020}



\end{document}